\documentclass[10pt]{wlscirep}
\usepackage[utf8]{inputenc}
\usepackage[T1]{fontenc}
\usepackage[inkscapeformat=png]{svg}
\usepackage[displaymath, mathlines]{lineno}

\usepackage{xcolor}

\title{Stereoscopic Depth Perception Through Foliage}

\author[1]{Robert Kerschner}
\author[1]{Rakesh John Amala Arokia Nathan}
\author[2]{Rafal Mantiuk}
\author[1,*]{Oliver Bimber}

\affil[1]{Johannes Kepler University Linz, AT}
\affil[2]{University of Cambridge, UK}
\affil[*]{oliver.bimber@jku.at}

\begin{abstract}
Both humans and computational methods struggle to discriminate the depths of objects hidden beneath foliage. However, such discrimination becomes feasible when we combine computational optical synthetic aperture sensing with the human ability to fuse stereoscopic images. For object identification tasks, as required in search and rescue, wildlife observation, surveillance, and early wildfire detection, depth assists in differentiating true from false findings, such as people, animals, or vehicles vs. sun-heated patches at the ground level or in the tree crowns, or ground fires vs. tree trunks. We used video captured by a drone above dense woodland to test users' ability to discriminate depth. We found that this is impossible when viewing monoscopic video and relying on motion parallax. The same was true with stereoscopic video because of the occlusions caused by foliage. However, when synthetic aperture sensing was used to reduce occlusions and disparity-scaled stereoscopic video was presented, whereas computational (stereoscopic matching) methods were unsuccessful, human observers successfully discriminated depth. This shows the potential of systems which exploit the synergy between computational methods and human vision to perform tasks that neither can perform alone.            
\end{abstract}
\begin{document}
\flushbottom
\maketitle
%
%
\thispagestyle{empty}
\section*{Introduction}
\label{sect:Introduction}

\begin{figure}[ht]
\centering
\includegraphics[width=\linewidth]{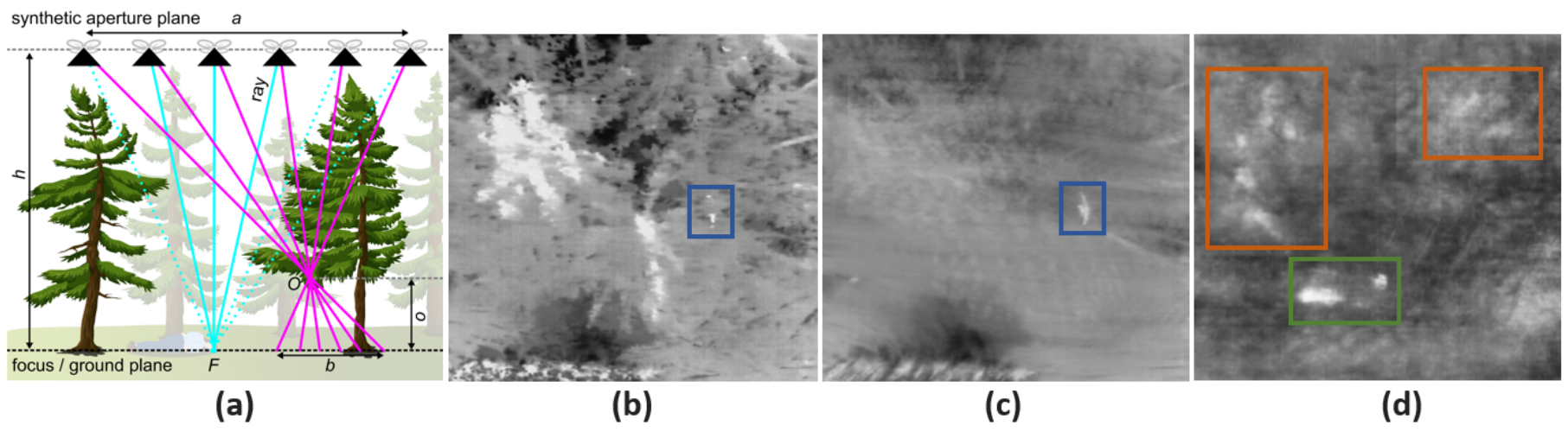}
\caption{Optical synthetic aperture sensing principle (a): Registering and integrating multiple images captured along a synthetic aperture of size \textit{a} while computationally focusing on focal plane  \textit{F} at distance \textit{h} will defocus occluders \textit{O} at distance \textit{o} from  \textit{F} (with a point-spread of  \textit{b}) while focusing targets on \textit{F}. Conventional thermal aerial image (b) of woodland with an occluded person on the ground (blue box). Suppressed occlusion when integrating 30 thermal images captured along a synthetic aperture of \textit{a}=14\,m at \textit{h}=26\,m AGL (c). An ambiguous example of an integral image in which true (lying and standing persons in the green box) and false (heated ground patches in red boxes) detections can be made. They cannot be differentiated by other discriminators, such as shape.}
\label{fig:AOS}
\end{figure}

Occlusion caused by vegetation, such as woodland, is a serious problem in aerial imaging tasks with drones or manned aircraft in applications including search and rescue, wildfire detection, wildlife observation, security, and surveillance. As an example, a thermal drone recording of a person occluded by a trees is shown in Fig. \ref{fig:AOS}b. 
The problem of occlusion can be reduced by the principle of  synthetic aperture sensing \cite{kurmi2018airborne,bimber2019synthetic,kurmi2019statistical,kurmi2019thermal,schedl2020airborne,kurmi2020fast,kurmi2021pose,schedl2020search,schedl2021autonomous,ortner2021acceleration,kurmi2022combined,amala2022through,seits2022role,amala2022inverse,seits2022evaluation,amala2023drone,nathan2023synthetic} (cf. Fig. \ref{fig:AOS}) -- a signal processing technique that computationally combines multiple measurements of sensors with physically limited apertures to mimic an advanced (virtual) sensor of a wider (synthetic) aperture with enhanced signal sensing capabilities. This method is applied to various signal types and sources, such as radar\cite{moreira2013tutorial,li2015synthetic,rosen2000synthetic}, radio telescopes\cite{levanda2010synthetic,dravins2015optical}, interferometric microscopy\cite{ralston2007interferometric}, sonar\cite{hayes2009synthetic,Hansen11}, ultrasound\cite{jensen2006synthetic,zhang2016synthetic}, LiDAR\cite{barber2014synthetic,turbide2017synthetic}, and optical imaging\cite{zhang2018synthetic,yang2016kinect,pei2019occluded}.

For optical synthetic aperture sensing, where image sensors with limited aperture optics are applied, a sequence of regular images with their telemetry data (e.g., captured by a drone while flying along a path whose length corresponds to an extremely wide synthetic aperture) are computationally combined as follows: 
The pixels from each camera image are projected onto a hypothetical (virtual) focal plane, which is at distance $h$ from the synthetic aperture's plane (i.e., at the altitude of the flight path), as shown in Fig.~\ref{fig:AOS}a. The pixels in each view that project to the same location on the focal plane are averaged. Aligning the focal plane with the forest floor and repeating this for all of its locations results in a shallow depth-of-field integral image of the ground surface (cf. Figs. \ref{fig:AOS}c,d).
It approximates the signal of a physically impossible optical lens of the size of the synthetic aperture. The optical signal of out-of-focus occluders, such as the tree crowns, is suppressed, while focused targets on or near the ground are emphasized. Computation of the integral images can be achieved in real time and is wavelength-independent. Thus, the method can be applied in the visible range, near-infrared range, or far-infrared range (thermal) to address many different use cases. It has previously been explored in search and rescue with autonomous drones\cite{schedl2020search,schedl2021autonomous}, bird census in ornithology\cite{schedl2020airborne} and through-foliage tracking for surveillance and wildlife observation\cite{amala2022through,amala2022inverse}. 

The main limitation of optical synthetic aperture sensing is that its results can be ambiguous if true targets cannot be differentiated from false targets on the basis of clear features such as shape. An example of this is illustrated in Fig. \ref{fig:AOS}d where strong thermal signatures of multiple potential targets near the forest floor are visible. While some of them are the results of sun-heating, only two are the true thermal signatures of people. With two-dimensional information alone, a clear distinction is impossible. The height differences between people and the forest floor could serve as an additional discriminator, but it is lost during projection. A computational 3D reconstruction from the sampled multi-view aerial images or the corresponding integral images is impossible in the case of strong occlusion\cite{kurmi2018airborne}, as shown and explained in \textit{Appendix 1}. Airborne laser scanning, such as LiDAR, has clear advantages over image-based 3D reconstruction when it comes to partially occluded surfaces, but it also has clear limitations\cite{kurmi2018airborne}: First, it is not sensitive to the target's emitted or reflected wavelengths. Thus, far-infrared (thermal) signals, for instance, cannot be detected. Second, due to mechanical laser deflection and high processing requirements, scanning and occlusion filtering of high-resolution point clouds cannot be achieved. Detection of heat signatures at high speeds and resolutions is, however, a main requirenment of many applications.
\\
\\
In this article, we explore the capability of the human visual system to detect these height differences via disparity depth cues when disparity-scaled stereoscopic depth perception and optical synthetic aperture sensing are combined. These differences can then serve as  additional discriminators in identification and classification tasks.

Let us analyze under what conditions the visual system can fuse and discriminate depth differences between small and occluded targets, such as standing humans (up to 2\,m) occluded by tall trees (15--20\,m), seen from high altitudes (20--30\,m for drones flying above tree level). First, objects that differ much in height (e.g., tree crowns vs. targets on the ground) and are located  closely together in the image will result in large disparity gradients (disparity difference divided by the distance between two objects). If the disparity gradient exceeds the limit of human visual perception, diplopia \cite{BURT1980615} will result, making stereoscopic function impossible. Second, if objects are seen from relatively far distances, and their height difference is small, the disparity difference between them may fall below the stereo acuity limit \cite{deepa2019assessment,filippini2009limits,tyler1974depth}, which makes depth discrimination impossible. The latter problem can be addressed by enlarging (scaling) disparities by assuming large viewing baselines (e.g. much larger than a typical inter-ocular distance of 6.5\,cm).

\begin{figure}[ht]
\centering
\includegraphics[width=0.75\linewidth]{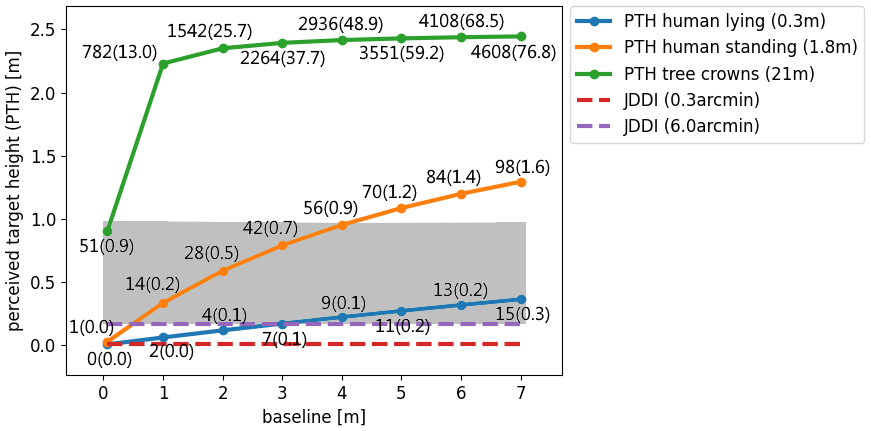}
\caption{The increase in perceived target height (PTH) with an increasing stereo baseline for three unoccluded objects of different heights (solid plots): tree crowns, lying person, and standing person. Stereo acuity sets the just-detectable depth interval (JDDI) required for perceiving height differences (dashed lines). Both the conservative (0.3\,acrmin) and the realistic (6\,arcmin) JDDIs are plotted. Disparities, or rather disparity gradients (numbers next to the markers), limit the maximum length of the baseline above which objects cannot be fused due to diplopia. Consequently, the grayed region represents the range in which depth can be perceived (assuming, for example, a disparity gradient limit of 1.0 and a stereo acuity of 6.0\,arcmin). Display disparities are given with respect to the ground level and for 60\,arcmin object distances. For this plot we assume the capturing and display parameters provided in the \textit{Methods} section.}
\label{fig:perception_plot_no_occlusion}
\end{figure}

Fig. \ref{fig:perception_plot_no_occlusion} illustrates these two problems for the unoccluded case. The perceived target height (PTH, y-axis) increases with an increased stereo baseline (x-axis). The solid lines show the increase in perceived target height for three different object types: tree crowns at 21\,m, a lying person at 0.3\,m, and a standing person at 1.8\,m above the surface. The numbers above the markers indicate the corresponding display disparities and disparity gradients for a given stereoscopic display (assuming minimal object distances of 60\,arcmin or 1\,deg). Depending on the stereo acuity, the just-detectable depth interval (JDDI) threshold changes (dashed lines). With poorer stereo acuity, larger depth intervals are required for perceiving height differences. Under these conditions and assuming an inter-ocular distance of 6.5\,cm, the height differences between target objects on the ground are unlikely to be detected  --- even if excellent stereo acuity is assumed. The disparity gradients of excessively large baselines and height differences, however, exceed the disparity gradient limit (e.g., 1 \cite{BURT1980615} - 3 \cite{MCKEE20021963}) and cannot be fused. How PTH and JDDI are determined is explained in \textit{Appendix 2}.  

\begin{figure}[ht]
\centering
\includegraphics[width=\linewidth]{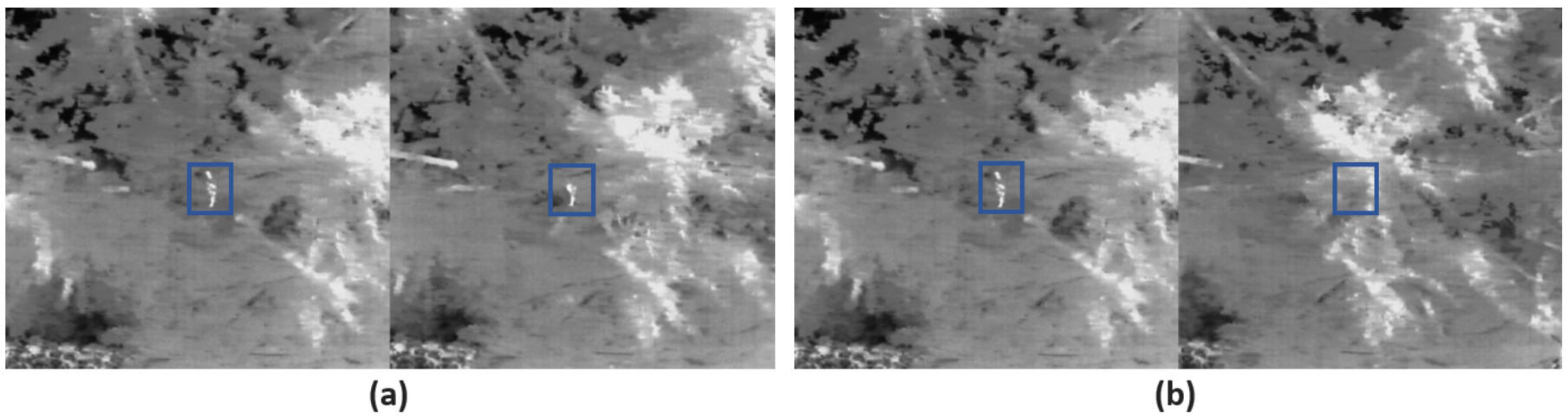}
\caption{Stereoscopic thermal aerial recordings of a sparsely occluded, person standing with arms outstretched to the sides (blue box) in woodland. View-dependent partial (a) or full (b) occlusion in stereo pairs cause binocular rivalry and prevent stereo fusion and consequently depth perception.}
\label{fig:binocular_rivalry}
\end{figure}

The third problem is that view-dependent occlusion in the stereo pairs causes binocular rivalry (which appears when radically different images are presented to each eye), which also prevents stereoscopic fusion if it is too strong \cite{treisman1962binocular,wilcox2007depth}. Examples are illustrated in Fig. \ref{fig:binocular_rivalry}. It has been found that, if partially occluded object fragments are horizontally aligned and match a continuous surface, our visual system tends to extrapolate a coherent surface at an incorrect depth \cite{FORTE20021225}. Horizontally aligned continuous object surfaces, however, are usually not present under realistic occlusion conditions such as ours. 
Although depth cannot be reconstructed computationally, we show that surface continuity can be reconstructed computationally, and that this enables human depth perception.                     

\begin{figure}[ht]
\centering
\includegraphics[width=\linewidth]{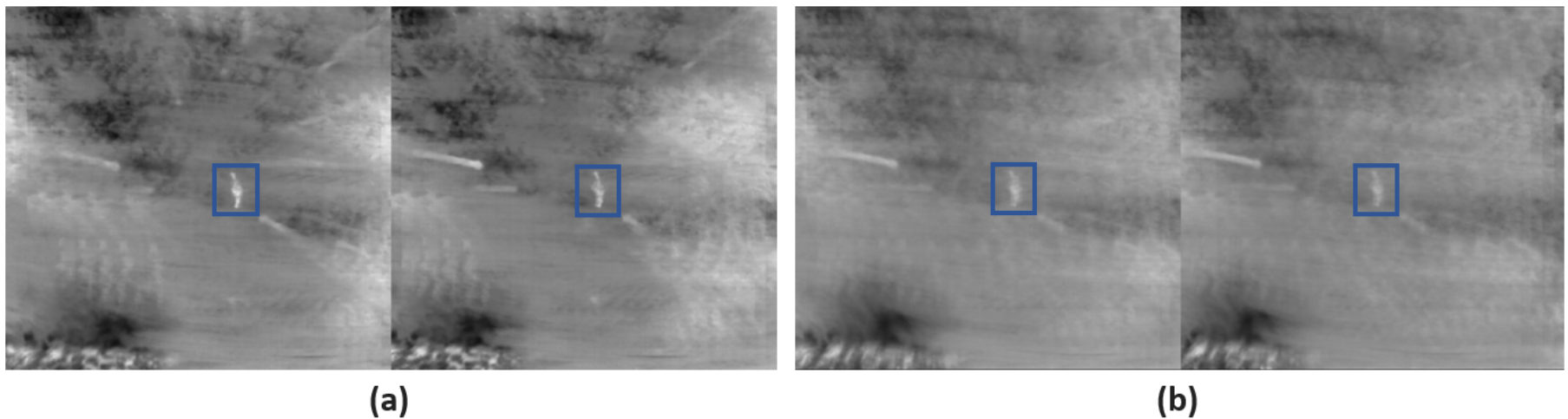}
\caption{Integral stereo pairs of the scenario shown in Fig. \ref{fig:binocular_rivalry}, where the synthetic aperture \textit{a} applied is smaller (a) or wider (b). The larger \textit{a}, the more shallow the depth of field. This leads to a reduction in sharpness and contrast.}
\label{fig:AOS_loss_contrast_frequencies}
\end{figure}

In this approach, we suppress occlusion by means of optical synthetic aperture sensing, as explained above. This also implies that we can compute stereoscopic integral images with suppressed occlusions for a given synthetic aperture of size \textit{a} and for two different viewing positions within \textit{a} and separated by a given baseline \textit{d}. With a large \textit{d} (larger than inter-ocular distance), we upscale disparities so they do not fall below the limits of stereo acuity. The larger \textit{a}, the more occlusion is suppressed, and binocular rivalry and extreme disparity gradients caused by tree crowns can consequently be reduced. However, a wide synthetic aperture also leads to a shallow depth of field and thus to defocus blur and lower contrast. The reduction in contrast and the loss of high spatial frequencies result in degradation of stereo acuity \cite{halpern1988contrast}. This is illustrated in Fig.~\ref{fig:AOS_loss_contrast_frequencies}.

With the results presented below, we make three main findings: First, occlusion removal in stereoscopic images is of fundamental importance for object identification tasks, while stereoscopic perception alone leads to no significant improvement. Second, while discriminating depth computationally (e.g., using 3D reconstruction from sampled multi-view images) is impossible in the case of strong occlusion, it becomes feasible visually by fusing binocular images with scaled disparities which result from optical synthetic aperture sensing. In fact, our observers were unable to discriminate depth with monoscopic viewing (even in the simplest case of unoccluded environments) and performed insignificantly better when viewing conventional stereo pairs. Not even motion parallax could enable them to determine depth differences. Third, the sampling and visualization parameters (best baseline and synthetic aperture size), although restricted for the reasons explained above, were found to be fairly consistent across all test cases evaluated.

Our findings demonstrate that it is possible to discriminate the depths of objects seen through foliage on the basis of optical synthetic aperture imagery captured with first-person-view (FPV) controlled drones or a manned aircraft. It has the potential to support challenging search and detection tasks in which occlusion caused by vegetation is currently the limiting factor. This includes use cases such as search and rescue, wildfire detection, wildlife observation, security, and surveillance. 
\section*{Results}
\begin{figure}[ht]
\centering
\includegraphics[width=\linewidth]{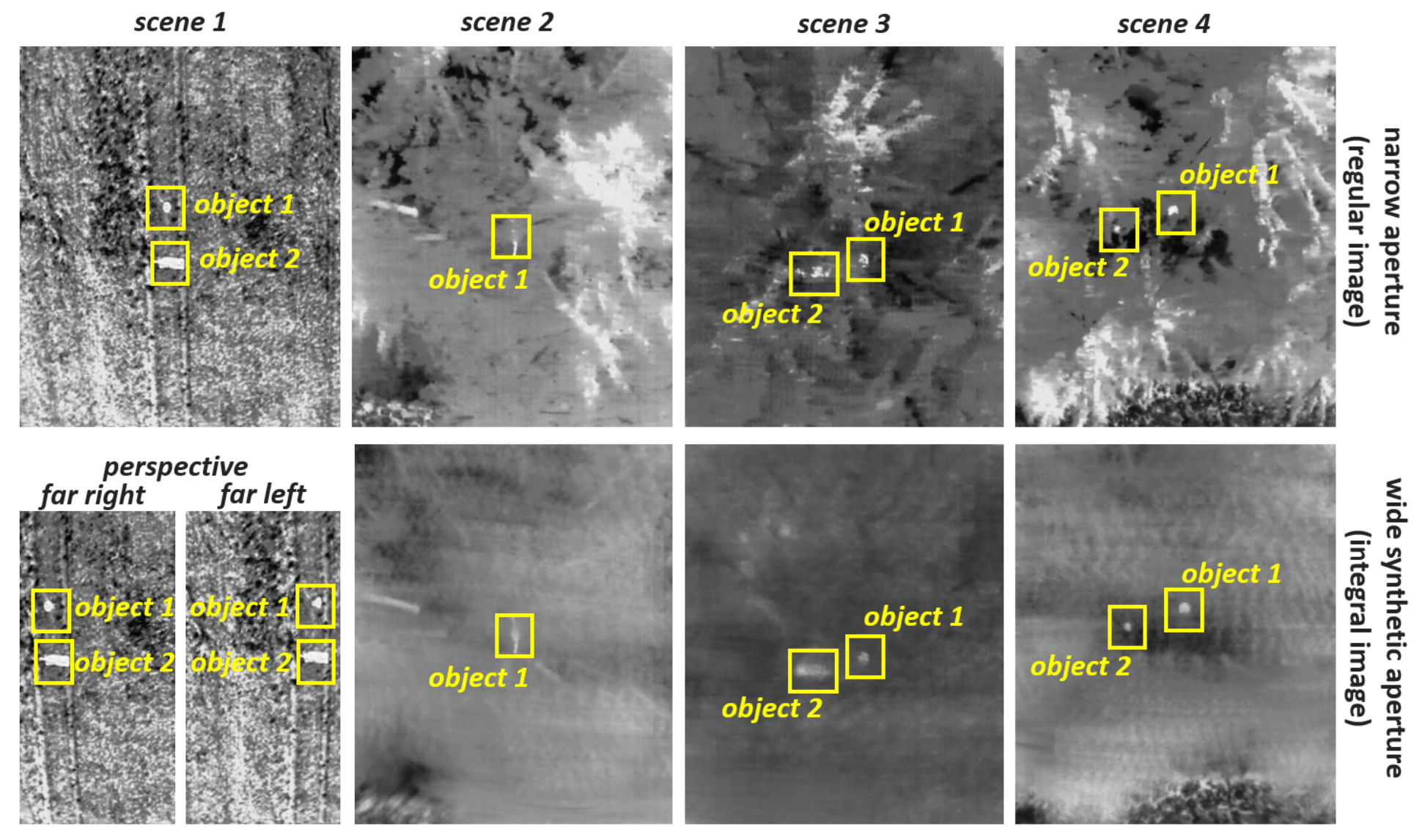}
\caption{Four test scenes with different occluded (\textit{scene 1}) and unoccluded (\textit{scenes 2-4}) target objects of various heights. A wide synthetic aperture suppresses occlusion in the resulting integral images. Changing horizontal perspective allows effects of motion parallax and less occluded viewing directions to be explored.}
\label{fig:scenes}
\end{figure}

In the course of experiments both in the field and subsequently with users, we tested the feasibility of relying on stereoscopic depth perception to identify objects and discriminate their depth seen through foliage in thermal recordings captured by a drone at 26\,m above ground level (AGL). The experiments were conducted for four different scenes (cf. Fig. \ref{fig:scenes}): an open field without vegetation (\textit{scene 1}) with a standing (\textit{object 1}) and a lying (\textit{object 2}) person; a forest (\textit{scene 2}) with one easily detected (based on shape features) person (\textit{object 1}) standing with arms outstretched to the sides; a denser forest (\textit{scene 3}) with a standing (\textit{object 1}) and a lying (\textit{object 2}) person; and a sparser forest (\textit{scene 4}) with a standing person (\textit{object 1}) and a 30\,cm high (roughly the height of a lying person) artificial object (\textit{object 2}) of similar shape, footprint, and temperature as the standing person.    

These recordings were then computationally combined to integral images (as explained in the \textit{Introduction}, cf. Fig. \ref{fig:AOS}a) and presented to 21 observers (11 female, 10 male, average age: 36) via a head-mounted stereoscopic display. At all times, the observers were able to use a game controller to interactively change the horizontal viewing perspective within the limits of the synthetic aperture $a$ covered. This allowed them to explore the effects of motion parallax and to find less occluded viewing directions. We changed visualization parameters, such as aperture size $a$ for integration and (camera) baseline $e_f$ for stereo-pair computation while asking the observers to describe the quality of the perceived images. The synthetic focal plane was set to the depth of the forest floor. Details on how the field and user experiments were carried out are provided in the \textit{Methods} section. 

\begin{figure}[ht]
\centering
\includegraphics[width=0.715\linewidth]{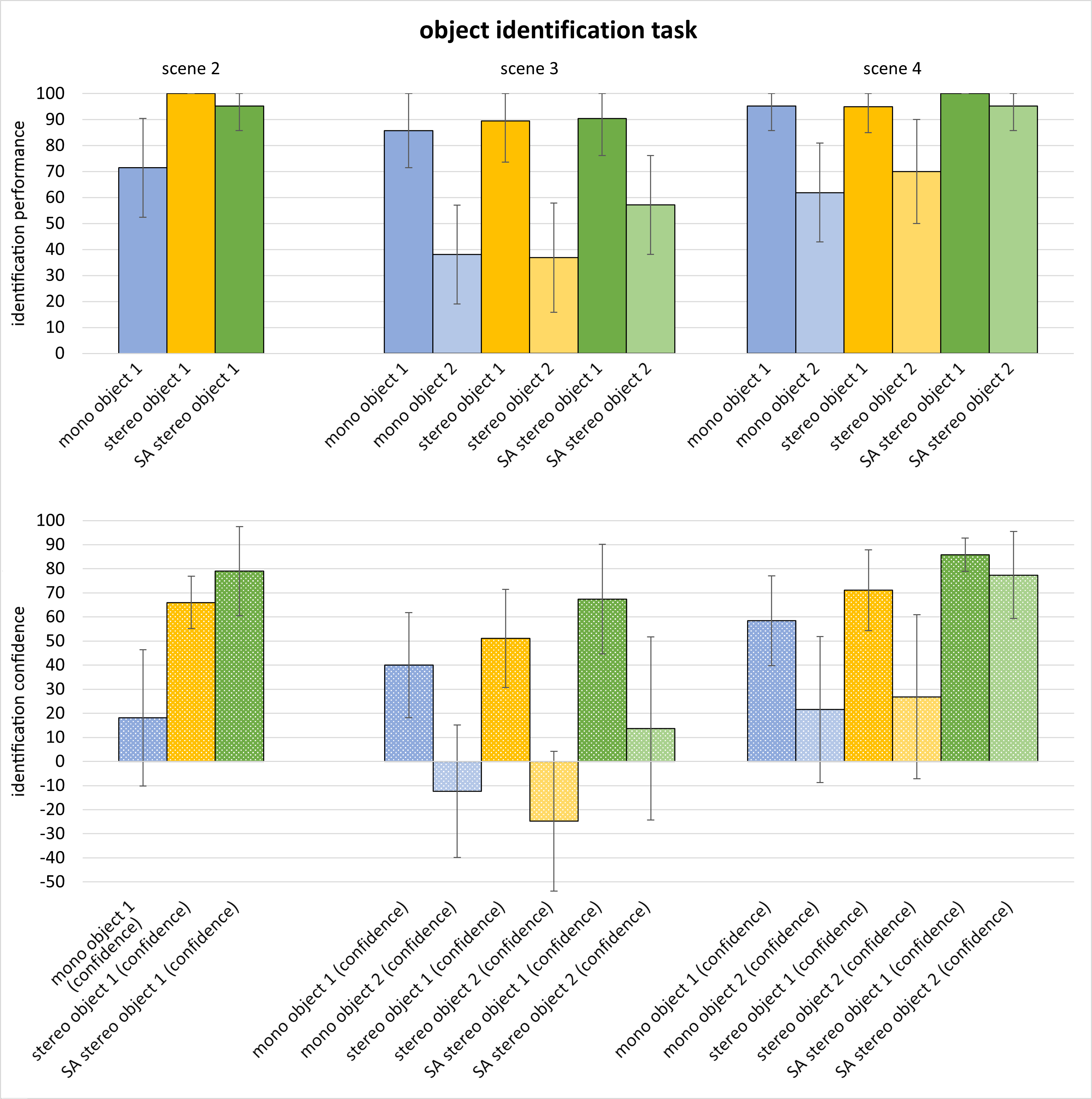}
\caption{Object identification performance (solid bars) and confidence values (dotted bars) for test scenes with occlusion (\textit{scenes 2-4}). We measured how often and with what confidence our target objects (\textit{object 1} and \textit{object 2}) were detected among all identifications and across all observers. The error bars denote 95\%
confidence intervals (based on the binomial distribution for performance and normal for observers' confidence).}
\label{fig:object_identification_task}
\end{figure}

In the first task (\textit{object identification}, cf. Fig. \ref{fig:object_identification_task}), we asked the observers to identify (i.e., to detect --- not to classify) all objects that appeared to be higher than the forest floor in our \textit{scenes 2-4} and to report on how confident they were on a scale from 0 (not confident at all) to 10 (very confident) in their decisions. Since the detection of unoccluded objects is trivial, \textit{scene 1} was skipped in this experiment. This task was repeated for regular monoscopic images without disparity (\textit{mono}), for regular stereoscopic images at different baselines $e_f$ (\textit{stereo}), monoscopic integral images at different synthetic apertures $a$ (\textit{SA mono}), and stereoscopic integral images at different synthetic apertures $a$ and different baselines $e_f$ (\textit{SA stereo}).   

Fig. \ref{fig:object_identification_task} illustrates how often and with what level of confidence our target objects (\textit{objects 1} and \textit{2}) were detected among all identifications. See the supplementary material for details on which objects were identified. Note that confidence values were counted as negative if our target objects were not identified. Thus, the negative values indicate misplaced confidence in identifications.   



We found no increase in identification performance when regular stereoscopic images were presented. However, identification performance and confidence improved (in particular for the more difficult to detect lower \textit{object 2}) when occlusion was removed in the stereoscopic images. This indicates that mainly occlusion removal leads to better detection of objects per se, and consequently in an increase of true identifications. False identifications, however, are still likely in case of missing distinctive features. If depth discrimination were available, false identifications might be reduced. Since our object identification experiment was a detection and not a classification task, true and false positive rates cannot be determined.  

\begin{figure}[ht]
\centering
\includegraphics[width=\linewidth]{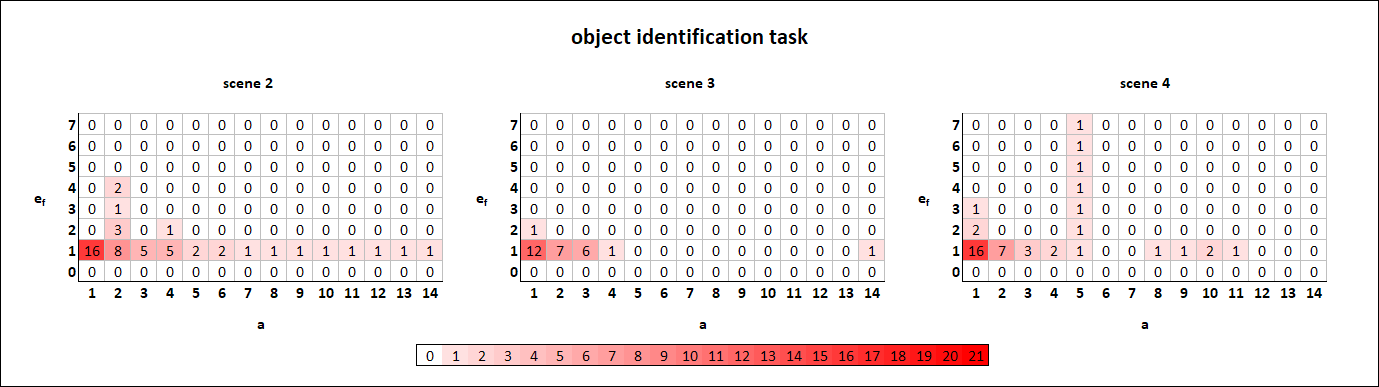}
\caption{Ranges of baselines $e_f$ and synthetic apertures $a$ for which our observers performed best in the object identification task. Note that observers performed best over a range of parameters rather than for exactly one pair. The color-coded numbers indicate the count of overlapping ranges across all observers (i.e., how often a parameter pair was considered best). The units of $e_f$ and $a$ are in meters. }
\label{fig:object_identification_task_heatmap}
\end{figure}

Figure \ref{fig:object_identification_task_heatmap} illustrates the ranges of baselines $e_f$ and synthetic apertures $a$ for which our observers performed best. Consistently across all three scenes, $e_f$=1--2\,m and $a$=1--4\,m were found to be optimal. Note that practical GPS positioning was too imprecise for sampling these parameters below 1\,m.   

\begin{figure}[ht]
\centering
\includegraphics[width=0.615\linewidth]{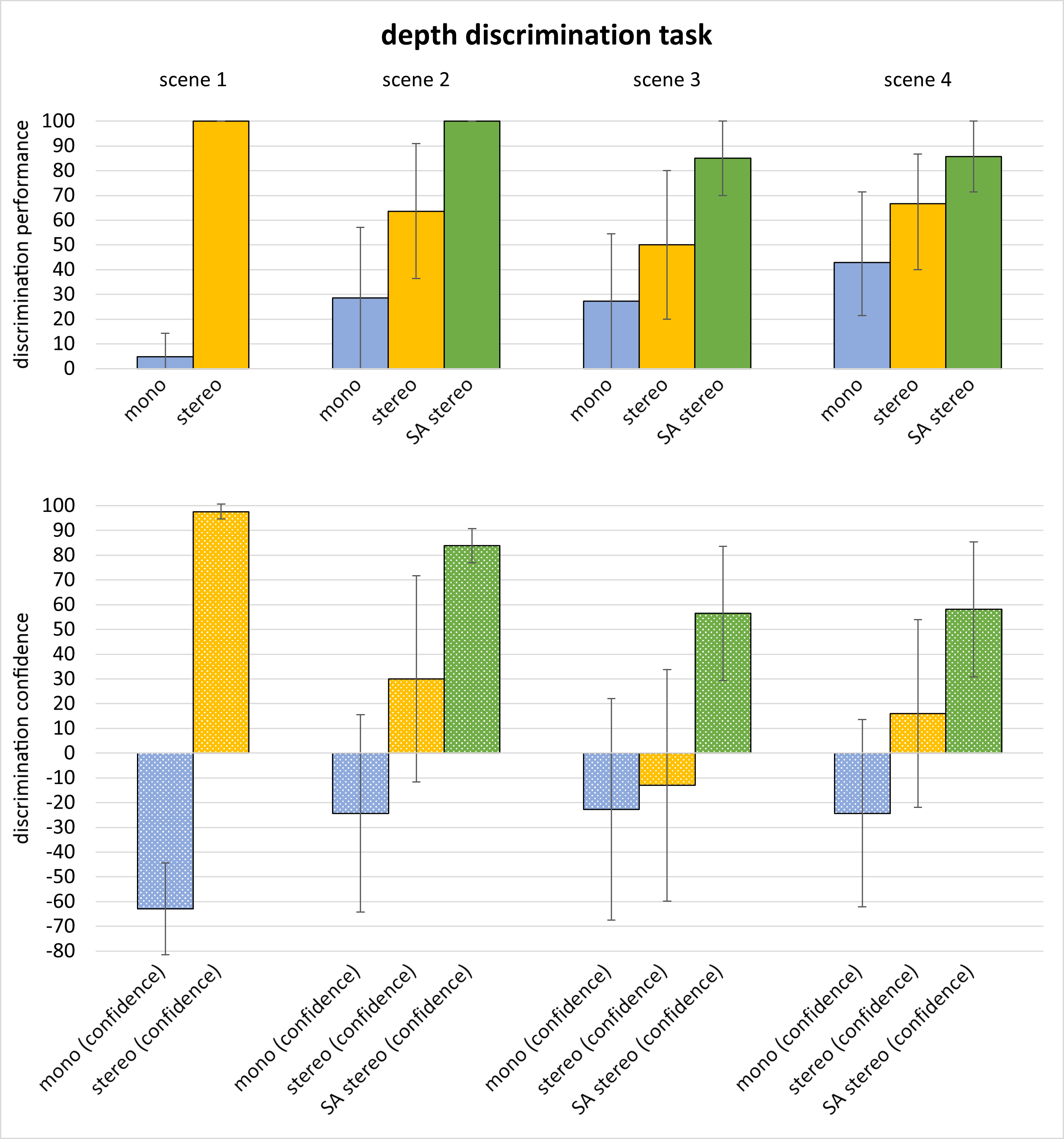}
\caption{Depth discrimination performance and confidence for all test scenes. We measured how often and with what level of confidence the true highest object (\textit{object 1}) was the highest object identified. The error bars denote 95\% confidence intervals (based on the binomial distribution for performance and normal for observers' confidence).}
\label{fig:depth_discrimination_task}
\end{figure}

In the second task (\textit{depth discrimination}, cf. Fig. \ref{fig:depth_discrimination_task}), we asked observers to indicate the highest object (which is always the standing person, \textit{object 1}, in our experiments) of those identified and again report how confident they were in their decisions. Here, we considered all four test scenes and varied $a$ for scenes with occlusion (\textit{scenes 2-4}) and $e_f$ for all scenes. The multi-view depth reconstruction results in \textit{Appendix 1} reveal that computational depth discrimination is infeasible for our occluded scenes.  

Our observers were also unable to discriminate depth in monoscopic images, even for the simplest case that is, without occlusion (\textit{scene 1}). Not even motion parallax was sufficient to enable them to determine depth differences. We assume that the monoscopic and parallax depth cues were simply too subtle for aerial viewing conditions (see also the most extreme horizontal perspectives of \textit{scene 1} in Fig. \ref{fig:scenes}). As soon as stereoscopic viewing was enabled and in case of no occlusion (\textit{scene 1}), all observers could discriminate depth with good accuracy. In the case of occlusion (\textit{scenes 2-4}), we observed a consistent marginal improvement in performance  when stereoscopic viewing was supported. The persistently low confidence scores, however, underline a remaining strong uncertainty. A significant improvement in both performance and confidence was observed when stereoscopic viewing was used together with the synthetic aperture sensing.    

\begin{figure}[ht]
\centering
\includegraphics[width=\linewidth]{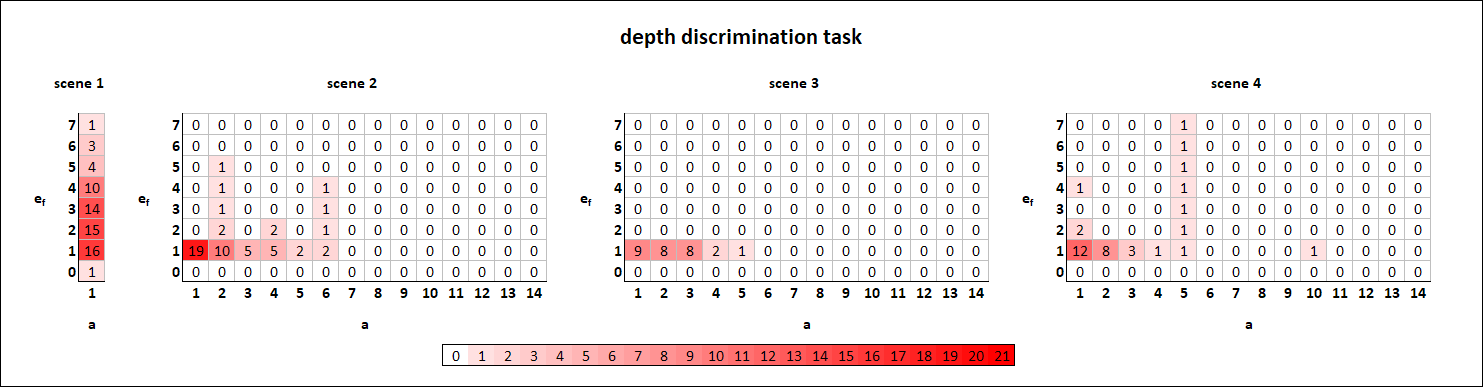}
\caption{Ranges of baselines $e_f$ and synthetic apertures $a$ for which our observers performed best in the depth discrimination task. Note that observers performed best within a range of parameters rather than in exactly one pair. The color-coded numbers indicate the count of overlapping ranges of all observers (i.e., how often a parameter pair was considered best). Note also that for the unoccluded scene (\textit{scene 1}) only $e_f$ is considered as a synthetic aperture for occlusion removal is not required. All units are in meters.}
\label{fig:depth_discrimination_task_heatmap}
\end{figure}

Figure \ref{fig:depth_discrimination_task_heatmap} illustrates the ranges of baselines $e_f$ and synthetic apertures $a$ for which our observers performed best. Consistently across all occluded scenes and in line with the theory explained in Fig. \ref{fig:object_identification_task_heatmap}, we found $e_f$=1-2\,m and $a$=1-4\,m to be optimal. For the case without occlusion (\textit{scene 1}), however, the average of best baselines was significantly higher than for the cases with occlusion. That the baselines were smaller in the scenes with occlusion can be related to the varying occlusion densities in a forest: For smaller baselines, the occlusions observed in the left and right views are similar, while they are more likely to vary more for larger baselines. 

\begin{figure}[ht]
\centering
\includegraphics[width=\linewidth]{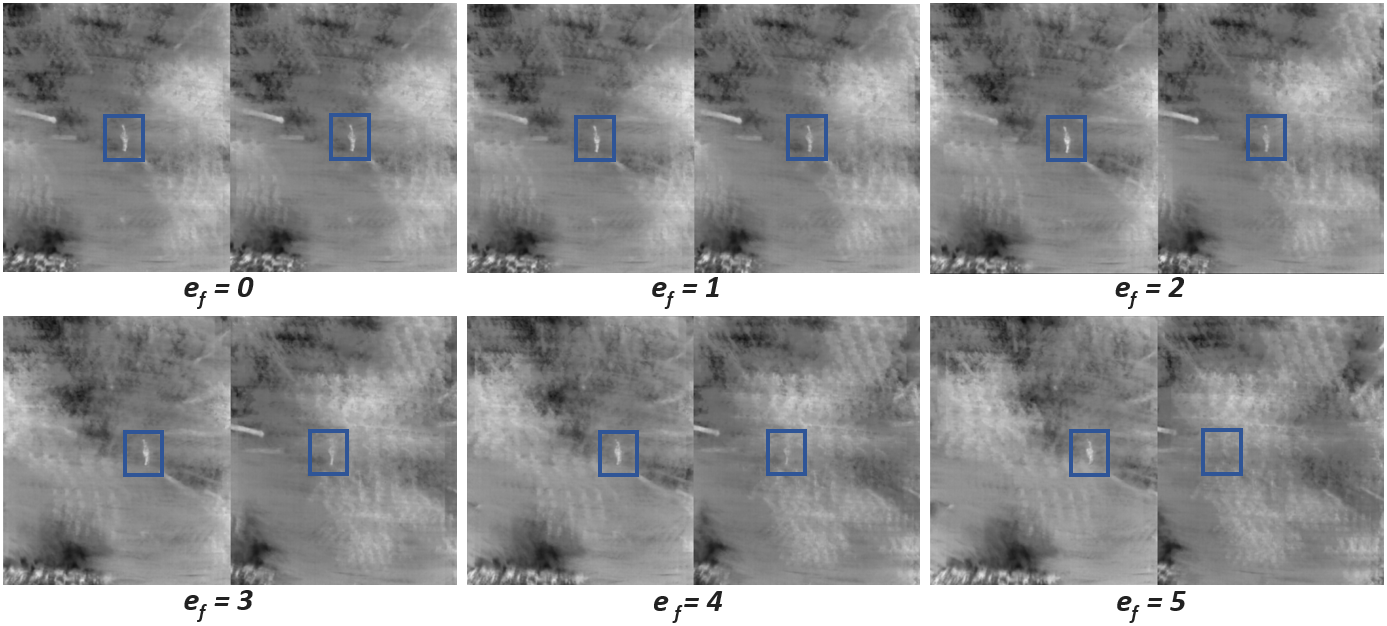}
\caption{Increasing binocular rivalry in integral stereo pairs with wider baseline due to locally varying occlusion density ($a$=2, in this example). All units are in meters.}
\label{fig:bin_riv_d}
\end{figure}

The occlusion differences between the left and right views cannot be compensated for with the same synthetic aperture $a$ for both eyes. Remaining occlusion that appears stronger in one view than in the other leads to binocular rivalry that negatively affects depth perception (cf. Fig. \ref{fig:bin_riv_d}). This is not the case for \textit{scene 1} without occlusion and the reason why the observers were able to increase the baseline up to the disparity gradient limit (cf. Fig. \ref{fig:perception_plot_no_occlusion}).     

Detailed results, the raw and the cleaned data, and the software systems used for our filed and user experiments are is available in the supplementary material (see \textit{Code and Data Availability}). 

\section*{Methods}
For our field experiments, we used a real-time kinematics (RTK) enabled DJI Mavic 3T drone with a 640$\times$512@30\,Hz thermal camera (61\,deg FOV, f/1.0, 5\,m-infinity focus). We developed a real-time imaging application using DJI’s Mobile SDK 5 that runs on the DJI RC Pro remote controller (Android 10). It supports real-time optical synthetic aperture scanning, occlusion removal, and interactive monoscopic and stereoscopic visualization. Visual results during flight can be presented either live on the remote controller's display (monoscopic) or on a head-mounted display attached to the remote controller's HMDI port (monoscopic or stereoscopic). See \textit{Code and Data Availability} section for how to obtain the imaging application. The data of our four test scenes was recorded at a constant altitude (26\,m AGL) by scanning a 1D synthetic aperture over a linear flight path of 14\,m and by choosing a sampling distance (0.5\,m) that was well above the GPS error. Consequently, a total of 29 thermal images were captured per scan. For a flying speed of 15\,m/s and an imaging speed of 30\,Hz, this takes approximately 1\,s, and the results are instantly displayed. As explained in the \textit{Introduction}, the scanned images are computationally combined to generate stereoscopic integral images, depending on the parameters chosen: synthetic aperture size $a$ (where $a$=14\,m is the maximum) and baseline $e_f$ (where $e_f$=14\,m-$a$ is the maximum). The synthetic focal plane was always kept on the ground (i.e., $h$=26\,m, cf. Fig. \ref{fig:AOS}a), since focus shifts of the order of the target heights (e.g., 0--2\,m AGL) did not significantly change the image content when observed from a large distance of 26\,m. Our test sites for \textit{scenes 2--4} were forested to various degrees with conifers and/or a variety of other tree species. The open field site for \textit{scene 1} was a freshly harvested corn field. 

All image data captured during the field experiments was recorded by the imaging application on the drone's remote controller and was used later in our offline user experiments. For the experiments, we  developed a visualization application that runs on desktop PCs or laptops (Microsoft Windows 11) and that reproduces the same visual experience as the imaging application during flight. It presents the image data to our observers via a head-mounted stereoscopic display without requiring them to be in the field during the actual scans. See \textit{Code and Data Availability} section for how to obtain the visualization application and the data used for our survey. The head-mounted stereoscopic display used was a 1920$\times$1024@60Hz Enmesi E 812 (68\,deg diagonal FOV, 2485.2\,mm focal distance, 152$\times$ magnification, 10\,mm eyebox, internally using two 2.1" 1600 x 1600 IPS microdisplays at up to 1058 PPI). Per-observer diopter settings were adjusted on the display before each session. A PowerA Nintendo Switch USB wired game controller was used to change the viewing perspective interactively. 

We tested a total of 21 observers (11 female, 10 male, average age: 36, the youngest 14, the oldest 67). Note that stereoscopic depth perception is considered to be fully developed at the age of 12\cite{nardini2010fusion}. As explained in the \textit{Results} section, we first performed the object identification task, then the depth discrimination task. On average, a survey round took a total of 45 minutes per participant. The order in which the scenes were presented matches that (left to right) shown in Figs. \ref{fig:object_identification_task} and \ref{fig:depth_discrimination_task}. In between the scenes, we displayed a neutral stereo pair to set stereoscopic fusion back to the same initial condition. While feedback from the participants was recorded in a questionnaire, all adjusted parameters were automatically recorded and stored by our application. To explore optimal visualization parameters, we incrementally increased $a$ and $e_f$ (starting with $a$=0 and $e_f$=0) while repeating each experimental trial of each task (object identification or depth discrimination) for each scene until the depth perception reported deteriorated. If both parameters were to be changed, we always started with $a$, followed by $e_f$. Participants were, at all times, able to interactively change their viewing positions using a game controller, and the time for stereo fusion was always allowed.         

\section*{Summary and Conclusion}
Identification and classification of objects that are strongly occluded by vegetation is aided significantly by the ability to discriminate their depths, which
provides important additional information to tell true from false findings, for instance, people, animals and vehicles from sun-heated patches of open ground or the tree crowns, or ground fires from tree trunks. This cannot be accomplished with conventional monoscopic or multi-view aerial images --- neither computationally nor by visual inspection of images or video. 

While neither human nor computer vision can perform this task on its own, we show that the synergy of both makes it possible. We have demonstrated this based on three main findings: First, occlusion removal in stereoscopic images is of dominant importance for object identification tasks while stereoscopic perception alone leads to no significant improvement. 
Second, depth discrimination significantly benefits from both removed occlusions and stereoscopic images with enlarged disparities. In fact, when our observers viewed the monoscopic images, they were unable to discriminate depth, even for the simplest case of unoccluded environments. Their performance was also poor when viewing conventional stereo pairs with occlusions, and even motion parallax could not provide sufficient hints for depth differences to be determined. Our third finding is that the relevant sampling and visualization parameters (best camera baseline and synthetic aperture size) are restricted for the reasons explained in the \textit{Introduction}, but were found to be fairly consistent throughout all test cases evaluated and in line with the theory explained in Fig. \ref{fig:perception_plot_no_occlusion}. 

While detecting depth differences computationally (e.g., using advanced multi-view 3D reconstruction) is impossible in the case of strong occlusion\cite{kurmi2018airborne} (see \textit{Appendix 1}), we have shown that the human visual system can perform this task robustly. One reason for this might be our visual system's ability --despite binocular rivalry-- to integrate partially occluded surfaces that appear sufficiently continuous in a horizontal viewing direction \cite{FORTE20021225}. While this continuity is not given in conventional stereo pairs with dense occluders, it is enhanced in stereoscopic integral images with wider synthetic apertures that suppress occlusion. We believe that this is the main reason why depth discrimination was significantly improved in stereo integral images, even though binocular rivalry was not fully eliminated in cases of locally varying occlusion densities and too wide baselines. 

Wider synthetic apertures reduce not only occlusions, but also lower image contrast (by blending multiple reprojected images). Therefore, for very large apertures, stereo acuity is decreased because of low contrast (i.e., the grayed region shown in Fig.~\ref{fig:perception_plot_no_occlusion} gradually shrinks towards the top). We can also deduce the range of effective baselines: While, if a baseline is too small, a disparity that falls below the just-detectable depth interval (JDDI in Fig.~\ref{fig:perception_plot_no_occlusion}) results, if the baseline is too large, the disparity gradient exceeds the limit for binocular fusion (upper edge of the gray box in Fig.~\ref{fig:perception_plot_no_occlusion}). All of these factors limit the usable imaging and visualization parameter ranges. 

In our experiments, the observers could discriminate disparity of 12\,arcmin (resulting from a height difference between a standing and a lying person from 26\,m viewing distance and at 1\,m baseline, see Fig. \ref{fig:perception_plot_no_occlusion}) seen through foliage. The relationship between depth discrimination precision, occlusion density, imaging, and display parameters has yet to be investigated. 
Furthermore, the results in Figs. \ref{fig:object_identification_task_heatmap} and \ref{fig:depth_discrimination_task_heatmap} suggest to explore apertures and baselines below 1\,m. Since GPS positioning is currently to imprecise in practice, this can be investigated in simulations.     

Deep-learning-based image restoration methods \cite{zhang2021plug} can potentially compensate for the contrast and sharpness loss in the integral images. It may also be possible to retain  stereo acuity with more advanced sampling: If a drone were equipped with two cameras at the optimum baseline distance ($e_f{\approx}1$\,m) and the video were captured in the direction orthogonal to the baseline, the defocus due to the synthetic aperture would affect contrast  only in the vertical direction, which is less relevant to binocular fusion. 
Such avenues should form part of future work. 

Our findings demonstrate that human operators can detect depth differences between objects seen through foliage with first-person-view (FPV) drones or manned aircraft, where the  thermal images captured are processed in real time with synthetic aperture sensing methods. This has the potential to support challenging search operations where occlusion caused by vegetation is currently the main limiting factor, as is the case for search and rescue, wildfire detection, wildlife observation, security, and surveillance.  

\section*{Appendix 1}
\begin{figure}[ht]
\centering
\includegraphics[width=\linewidth]{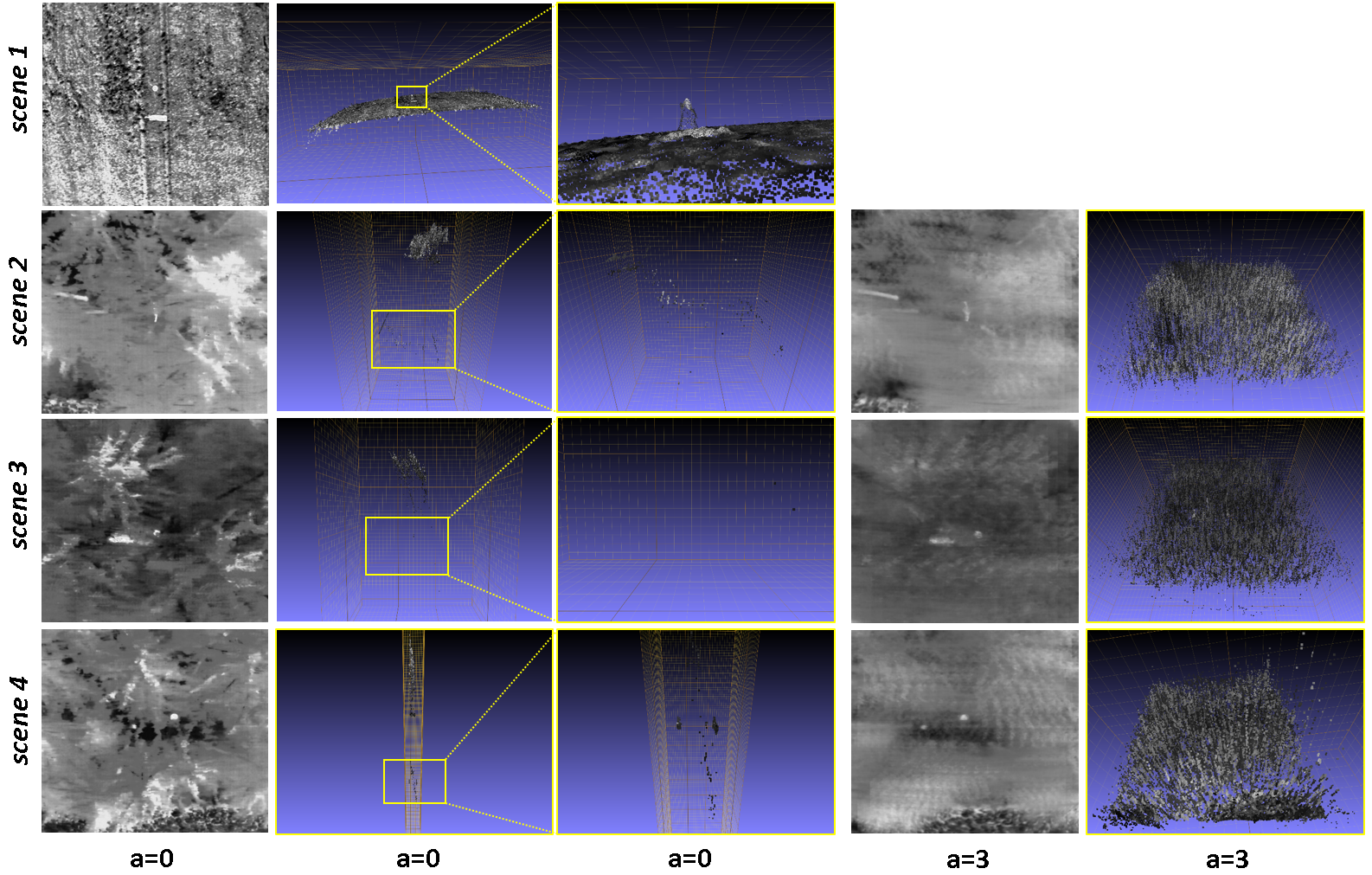}
\caption{Multi-view 3D reconstruction results of our four test scenes. With regular aerial images ($a=0$) only the upper depth layer can be reconstructed. For the unoccluded \textit{scene 1}, it contains the two targets on the ground. For all occluded scenes, it contains --at best-- the tree crowns. Yellow boxes show close-ups of of the reconstructed ground surface. If occlusion-suppressed integral images (with $a$=3 in this example) are used instead of regular aerial images, targets remain undetectable in noisy reconstructions of the ground surface. All units are in meters.}
\label{fig:3D_reconstructions}
\end{figure}

Fig. \ref{fig:3D_reconstructions} illustrates 3D reconstruction results of our four test scenes, computed with the state-of-the-art structure-from-motion and multi-view stereo pipeline, COLMAP \cite{schoenberger2016mvs,schoenberger2016sfm}.  

While the occlusion-free \textit{scene 1} can be fully reconstructed from the recorded aerial images, and the height difference between the two targets (standing and lying person) can easily be determined computationally, depth estimation fails for all other scenes. Here, at best, only tree crowns can be partially reconstructed.  
Using integral images from multiple perspectives on the synthetic aperture does not reconstruct the occluding tree structures, but an extremely noisy ground on which the targets cannot be detected at all.

The reason for this behavior is that without occlusion removal only the unoccluded tree crowns provide consistent and strong image features over a sufficient number of perspectives. The appearance of image features of occluded objects below is too inconsistent to be matched properly. With optical synthetic aperture sensing, in contrast, occlusion caused by the tree crowns is suppressed and is therefore not reconstructed. This comes at the cost of image features of the remaining ground surface loosing contrast and high frequencies (sharpness) in general. They become insufficient for computational stereo-matching - but are obviously sufficient for perceptual stereo fusion.   

Note that for each scene all images captured were used and that 3D reconstruction took approx. 15\,min on a modern desktop computer.  
\section*{Appendix 2}
For a given screen distance $v$, inter-ocular eye distance $e$, and object disparity $d$, the perceived object distance $z$ is given by \cite{hainich2016displays}

\begin{equation}
z=\frac{ev}{e-d}.
\label{eqn:perceived_object distance}
\end{equation}

It follows that

\begin{equation} 
d=\frac{e(z-v)}{z}.
\label{eqn:disparity}
\end{equation}

Applying Eqn. \ref{eqn:disparity} to compute the disparity on the focal plane at distance $v_f$ (equals $h$ in Fig. \ref{fig:AOS}a), camera baseline $e_f$ on the synthetic aperture plane, and target distance $z_f=v_f-h_t$ ($h_t$ is the target height) from the synthetic aperture plane; and then scaling the resulting disparity to the display parameters to determine the perceived object distance on the display $z_d$ using Eqn. \ref{eqn:perceived_object distance} results in

\begin{equation} 
z_d=\frac{e_dv_d}{e_d-\frac{v_d\tan(FOV_d/2)e_f(z_f-v_f)}{v_f\tan(FOV_f/2)z_f}},
\label{eqn:perceived_object_height}
\end{equation}

where $e_d$ and $v_d$ are the inter-occular eye distance and the distance of the display image plane, and $FOV_d$ and $FOV_f$ are the fields of view of the display and camera, respectively. 

Consequently, the perceived target height is

\begin{equation} 
PTH=v_d-z_d.
\label{eqn:PTH}
\end{equation}

The just-detectable depth interval is given by 
 \cite{howard2002seeing,howard1919test}

\begin{equation} 
JDDI=\frac{d_\gamma v_d^2}{ce_d+v_d},
\label{eqn:JDDI}
\end{equation}

where $d_\gamma$ is the stereo acuity (in arcmin) and $c=3437.75$ (1 radian in arcmin). 

\newpage
\bibliography{refs}

\section*{Acknowledgements}

This research was funded by the Austrian Science Fund (FWF) and German Research Foundation (DFG) under grant numbers P32185-NBL and I 6046-N, as well as by the State of Upper Austria and the Austrian Federal Ministry of Education, Science and Research via the LIT-Linz Institute of Technology under grant number LIT2019-8-SEE114.
\section*{Author Contributions Statement}
O.B. developed the concept. R.K. and R.N. implemented the algorithms. R.K., R.N., R.M., and O.B. conceived the experiments, R.K. and R.N. conducted the experiments. R.K. and O.B. and R.M. analyzed the results. O.B., R.K., R.N., and R.M. wrote the paper. All authors reviewed the manuscript. 
\section*{Additional Information}

\textbf{Code and Data Availability:} The python script with the equations in Appendix 2 which was used for computing Fig. \ref{fig:perception_plot_no_occlusion}, the stereoscopic visualization application and image data used for our user experiments, the raw and cleaned results of our user experiments, and the 3D reconstructions shown Appendix 1 are available at https://doi.org/10.5281/zenodo.8423145.
Since the drone imaging application used for recording our data is a dual use technology, it is available on request from https://github.com/JKU-ICG/AOS/.  
\\
\textbf{Competing interests:} None.
\end{document}